# The Role of Artificial Intelligence (AI) in Adaptive eLearning System (AES) Content Formation: Risks and Opportunities involved.

Suleiman Adamu[1], Jamilu Awwalu[2]

[1]Information and Communications Technology Unit, Sule Lamido University Kafin Hausa, Jigawa State, Nigeria.
[2]Department of Computer Science, Faculty of Computing and Applied Sciences, Baze University Abuja, Nigeria.
[1]s.adamu@jsu.edu.ng, [2]jamilu.awwalu@bazeuniversity.edu.ng

## Abstract

Artificial Intelligence (AI) plays varying roles in supporting both existing and emerging technologies. In the area of Learning and Tutoring, it plays key role in Intelligent Tutoring Systems (ITS). The fusion of ITS with Adaptive Hypermedia and Multimedia (AHAM) form the backbone of Adaptive eLearning Systems (AES) which provides personalized experiences to learners. This experience is important because it facilitates the accurate delivery of the learning modules in specific to the learner's capacity and readiness. AES types vary, with Adaptive Web Based eLearning Systems (AWBES) being the popular type because of wider access offered by the web technology.The retrieval and aggregation ofcontents for any eLearning system is critical whichis determined by the relevance of learning material to the needs of the learner.In this paper, we discuss components of AES, role of AI in AES content aggregation, possible risks and available opportunities.

## Introduction

The Internet contains several categories of materials, some of which can be used as educational resources. These educational resources can be retrieved for content formation of Adaptive eLearning Systems (AES). Adaptive eLearning is a branch of eLearning that provides educational materials and resources based on the learner's needs. The adaptive learning system is not limited to delivering learning materials in a personalized manner to the learner, but it also extends to adapting in terms of interaction with learners, and maintains learner's preferences.

The three main characteristics an AES must have according to Vassileva (2012) are:

i. its behavior should be adaptable according to changes in the working environment or parts/components of the adaptive system itself
ii. constituent modules should be developed to maximize compliance with thechanges in external environments
iii. tools should be provided for monitoring and controlling its work, means for changing parameters and using a closed circle of actions to improveproductivity and to optimize the interaction with users

AES technology comprises Intelligent Tutoring System (ITS) and Adaptive Hypermedia and Multimedia (AHAM) as shown in figure 1.

Artificial Intelligence plays a key role in adaptive eLearning by providing personalized experience of learning. It plays important role specifically in the domains ontology and knowledge retrieval which are part of the main modules of the adaptive eLearning system. Retrieval and formation of correct contents for an eLearning system is critical. The success of any e-learning system is determined by the relevance of learning material to the needs of the learner.

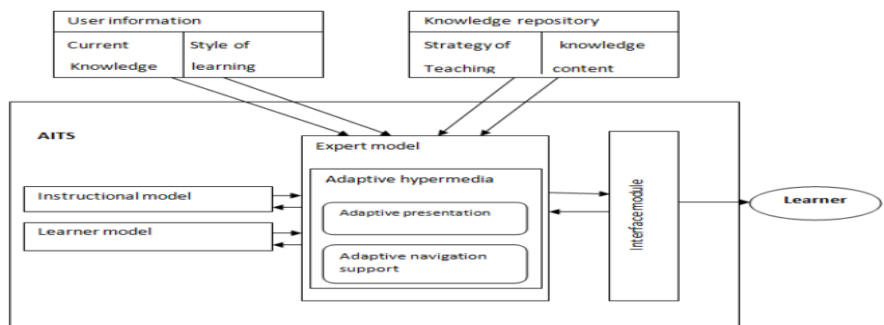

Figure 1 Adaptive Intelligent Tutoring System for eLearning. Source Sadakatullah (2014)

Unlike the traditional Learning Management Systems (LMS) where the instructor is responsible for the integrity of information passed unto the learners, the role Artificial Intelligence plays in knowledge retrieval is not and cannot be responsible for the information served to the leaner as algorithms are used for the knowledge retrieval, and these algorithms cannot be held responsible for information passed to learners as they serve the learners what they are able to retrieve and process.

**Components of an AES**

Several components as shown in figure 2 interact to make an AES. According to Potode & Manjare(2015), these components are the student model, teaching model, domain knowledge, communication module, expert modules, and the learning environment. However, the three models that make an AES according to Almohammadi et al.(2017) are the learners profile or model which is used to infer and profile the student's learning characteristics. According to Roy & Devshri(2011), the student model includes knowledge about personal traits, level of skills, and learning material access pattern. The taught content model, this contains modules to be learnt by the student. The instructional model matches how the learning contents are dynamically suggested to the learner. In particular, the success or failure of an AES heavily depends on the richness and correctness of the taught content module in terms of getting the right contents to the right leaner in the right order at the right time.

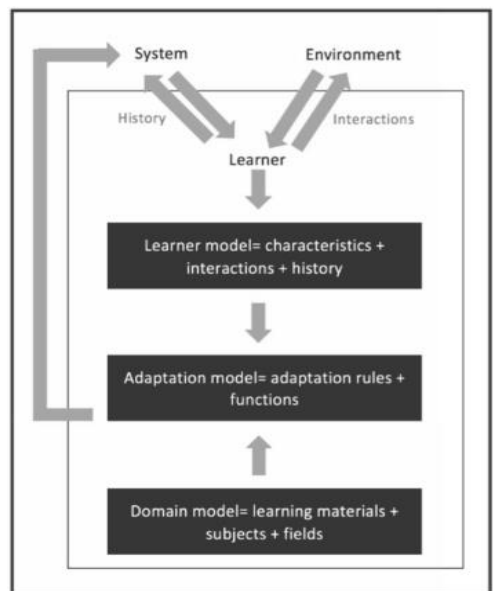

Figure 2 Components of Adaptive eLearning Systems. Ennouamani & Mahani(2017)

**Artificial Intelligence (AI) Techniques in Adaptive eLearning System (AES)**

Several researchon AES have been conducted with different approaches to improving content formation and delivery of AES. In the aspect of retrieving learning materials for content formation, Fouad(2012) proposed a fuzzy clustering system which analyses web log to extract leaners interested terms in visited pages. On the other hand, decision tree system was employed by Lin et al.(2013) for Personalized Creativity Learning System (PCLS). Also Baradwaj & Pal(2011) employed decision tree in extracting knowledge todescribestudents'performance which is vital when forming the student module of an AES. While Fuzzy Logic and Decision Tree are AI white box approach in AES because they can be easily checked and understood by an ordinary user and are suitable for adapting the dynamic nature of eLearning process Almohammadi et al.(2017), black box AI approaches such as Neural Networks byBeetham & Sharpe(2013) Khudhair & Ahmed(2017) and Bayesian Networks byGarcía(2005) Kardan, Speily, & Yosra(2015) have also been employed by researchers in AES.Sarasu & Thyagharajan(2015) combined black and white box AI approach in AES by using Adaptive Neuro Fuzzy Inference System (ANFIS) which combines neural network and fuzzy inference system.

**Risks and Opportunities**

Although AES offers the instructors the opportunity of authoring and engineering content, the level of content authoring is limited by the fact that the AI based AES must not be constrained to particular resources or repositories because it is autonomous and intelligent enough to search, retrieve, and update its knowledge base regularly. The risk involved here is that authoring in terms of obscene, derogative, and prohibited keywords filtering by instructors as they engineer and form contents, the AES may no filter such words. This is clear as seen on social media where such filtering has been enforced on obscene and derogative words, but users maneuver around by replacing a word or two with asterisk or exclamation marks such as *sh!tor sh_t*and *f\*ck or f_ck*. For a student whose medium of instruction is the English language, this can be a serious problem because he/she may mistake such words as normal words in a formal conversational or academic setting. Such derogative words were used by Microsoft AI twitter chat bot Tay where it even went on to claim Hitler was right and equate feminism with cancer.

AES systems that use the leaner's web log for content formation chances of not getting the appropriate learning content can be high. This is because according to internet usage statistics from SimilarWeb (2018), the top 10 websites with highest traffic worldwide are; search engines, arts and entertainment, social networks, adult, and encyclopedia. These according to classification are not mainly educational websites. This implies that majority of internet activities are not based on educational resources, and if such web log is to be used for content formation, then that could make a defective educational content.

Despite the risks involved in AI's role in AES content formation, a well designed and implemented AES offers different opportunities to both students and teachers. Students benefit from opportunities such as, adaptive presentation in terms of sequencing and navigation support preferences, peer help and collaborative learning, and learning by practice. Another opportunity for students is the explanation facility AI offers on how a correct answer is arrived at and what the learner actually missed or didn't do correctly, this is contrary to what is available on traditional Learning Management Systems (LMS) where the correct answer is shown without giving the learner the opportunity of querying how that answer was arrived at and what the learner missed exactly. Also, AI helps adaptive eLearning systems with wider reach to knowledge materials and multilingual learning service of the same knowledge contents without needs of translation.

**Conclusion**

As technology grows, most of the current risks associated with AI AES would certainly be reduced or eliminated because these risks form research areas and interests for AI researchers that are currently looking at how they can be solved. The more we are predisposed to technology, the more risks associated with it are identified, for certain today's challenges and risks would not persist in years to come, but different and new challenges that the research community would take over to explore how they can be solved.